\documentclass[10pt,twocolumn,letterpaper]{article}

\usepackage[pagenumbers]{wacv}

\usepackage[pagebackref,breaklinks,colorlinks,allcolors=wacvblue]{hyperref}
\usepackage{array}
\usepackage{float}                      
\usepackage{adjustbox}                  
\usepackage[table]{xcolor}              
\usepackage{booktabs}                   
\definecolor{wacvblue}{rgb}{0.21,0.49,0.74}
\definecolor{oursblue}{RGB}{214,228,248}  
\definecolor{oursblue}{RGB}{214,228,248}
\definecolor{deltagreen}{RGB}{0,140,70}
\definecolor{deltared}{RGB}{200,30,30}
\definecolor{lastrowbg}{HTML}{F5F5F8}
\usepackage[most]{tcolorbox}   

\newtcolorbox{promptbox}{
  breakable,
  colback=gray!5,
  colframe=black!60,
  boxrule=0.4pt,
  arc=2pt,
  left=6pt,
  right=6pt,
  top=5pt,
  bottom=5pt,
  fontupper=\footnotesize
}

\def\wacvPaperID{41}
\def\confName{WACV}
\def\confYear{2027}

\title{Oddin: Latent Visual Reasoning in Unified Multimodal Models \\ via Self-Distillation from Image Generation}

\author{
Jaeyong Ko$^{*}$ \qquad
Hankyeol Kim$^{*}$ \qquad
Pilsung Kang$^{\dagger}$\\[4pt]
{\normalsize
Department of Industrial Engineering, Seoul National University}\\
{\normalsize Seoul, Republic of Korea}\\[2pt]
{\small\ttfamily
\{jyko22, hankyeol, pilsung\_kang\}@snu.ac.kr}\\[4pt]
{\footnotesize
$^{*}$Equal contribution.
\qquad
$^{\dagger}$Corresponding author.}
}

\begin{document}
\maketitle
\begin{abstract}
Unified large multimodal models (ULMMs) are able to perform visual reasoning through interleaved text-image generation, but generating explicit intermediate images incurs significant computational overhead. While visual latent reasoning addresses this, most existing methods rely on vision-language models (VLMs) incapable of visual generation. We propose Oddin, a novel framework enabling latent visual reasoning in ULMMs. Oddin replaces intermediate images with compact latent tokens processed entirely by the understanding branch, reserving the generation branch for the final output image. We train the model via offline self-distillation with three objectives, final image generation losses, text distillation losses, and cross-entropy losses for special tokens. Oddin surpasses interleaved generation baselines and closed-source VLMs on average across seven VBVR-Pro visual reasoning tasks. Qualitative analysis reveals that latent tokens attend to relevant regions in an input image, demonstrating that ULMMs can perform complex visual reasoning without generating intermediate images.
\end{abstract}
    
\section{Introduction}
\label{sec:intro}

Unified large multimodal models (ULMMs) aim to understand and generate both text and images within a single architecture. Early approaches adopt a single shared transformer backbone to process both modalities \citep{Transfusion,ShowO}. However, since visual understanding and generation require distinct capabilities, fully sharing parameters leads to suboptimal performance \citep{Janus,BAGEL}. Recent models address this by introducing decoupled branches for understanding and generation \citep{BAGEL,SenseNova-U1,SenseNova-U1.5}. By operating on a shared multimodal sequence, these models can reason over images and generate new ones within a single context.

Building on this capability, recent studies apply ULMMs to visual reasoning through interleaved generation of text and images, showing that the resulting multimodal interleaved reasoning outperforms text-only Chain-of-Thought (CoT) \citep{CoT} on certain tasks \citep{MVoT,Zebra-CoT,ThinkMorph}. However, generating intermediate images incurs computational overhead, as the model must reconstruct pixel-level details, including redundant background regions. In contrast, the information passed between reasoning steps is far more compact.

Visual latent reasoning addresses this inefficiency by replacing explicit intermediate images with compact latent representations \citep{Mirage,Monet,LVR,CoVT,ILVR,DeepLatent,LatentSketchpad}. However, most existing methods are built on vision-language models (VLMs), which cannot generate images. As a result, they are inapplicable to visual reasoning tasks requiring visual outputs, such as VBVR-Pro \citep{VBVR-Pro}. ULMMs avoid this limitation, since their generation branch can produce images. This raises a natural question: can a ULMM reason with compact latents in place of intermediate images, while still generating the final visual output?

We propose Oddin,\footnote{Short for ``odd interleaved generation," and a nod to Odin, the Norse god who sacrificed one eye to gain wisdom.} a novel training framework for latent reasoning in ULMMs. Oddin replaces each intermediate image with a latent token block processed by the understanding branch, while the generation branch only produces the final image. We train Oddin via offline self-distillation, using reasoning traces generated by the model itself through interleaved generation. Without explicitly supervising the latent tokens with intermediate images, we train the latent tokens to implicitly acquire the required reasoning capabilities at their positions. To achieve this, we train \texttt{SenseNova-U1.5} \citep{SenseNova-U1.5} with three losses: a final image generation loss, a text distillation loss, and a special-token cross-entropy loss.

Oddin surpasses the performance of interleaved generation and other baselines.
Qualitative analysis reveals that the latent tokens attend to the specific input regions that the intermediate image was supposed to draw. These findings show that a unified model can perform visual reasoning within the understanding branch without relying on intermediate image generation, while reserving the generation branch for producing the final image.
\section{Related work}

\paragraph{Interleaved Generation for Visual Reasoning.}
For visual reasoning, ULMMs can perform interleaved generation, producing intermediate images within reasoning trace~\citep{MVoT,TwGI,Zebra-CoT,ThinkMorph}. Most datasets and benchmarks used in this line of work, such as V*~\citep{VStar} and Zebra-CoT~\citep{Zebra-CoT}, target questions with text answers. Recent ones such as VBVR-Pro~\citep{VBVR-Pro}, RIG-Bench~\citep{RIG-Bench}, and GenPuzzle~\cite{GenPuzzle} require the answer itself to be an image. Although intermediate images do not always help~\citep{AdaViG}, interleaved generation in recent ULMMs is now strong enough on suitable tasks~\citep{VBVR-Pro, SenseNova-U1} to serve as a teacher, which Oddin exploits through self-distillation.

\paragraph{Visual Latent Reasoning.}
Extending latent reasoning in language models~\citep{Coconut,ImplicitCoT}, visual latent reasoning replaces explicit intermediate images with compact latent tokens~\citep{Mirage,LVR,Monet,SkiLa,LIVR}. However, these methods are built on VLMs and always answer in text, so they cannot handle tasks that require image outputs. In unified models, latent reasoning has mainly been applied to text-to-image generation~\citep{MILR,LAC}. LatentUM~\citep{LatentUM} reasons over visual states in a shared semantic space, but each state is still a full sequence of generated visual tokens, and the model requires new pre-training. Oddin adapts a pretrained ULMM to interleaved latent visual reasoning through offline self-distillation, replacing each intermediate image with latent tokens.

\section{Method}

\begin{figure*}[t]
    \centering
    \includegraphics[width=\textwidth]{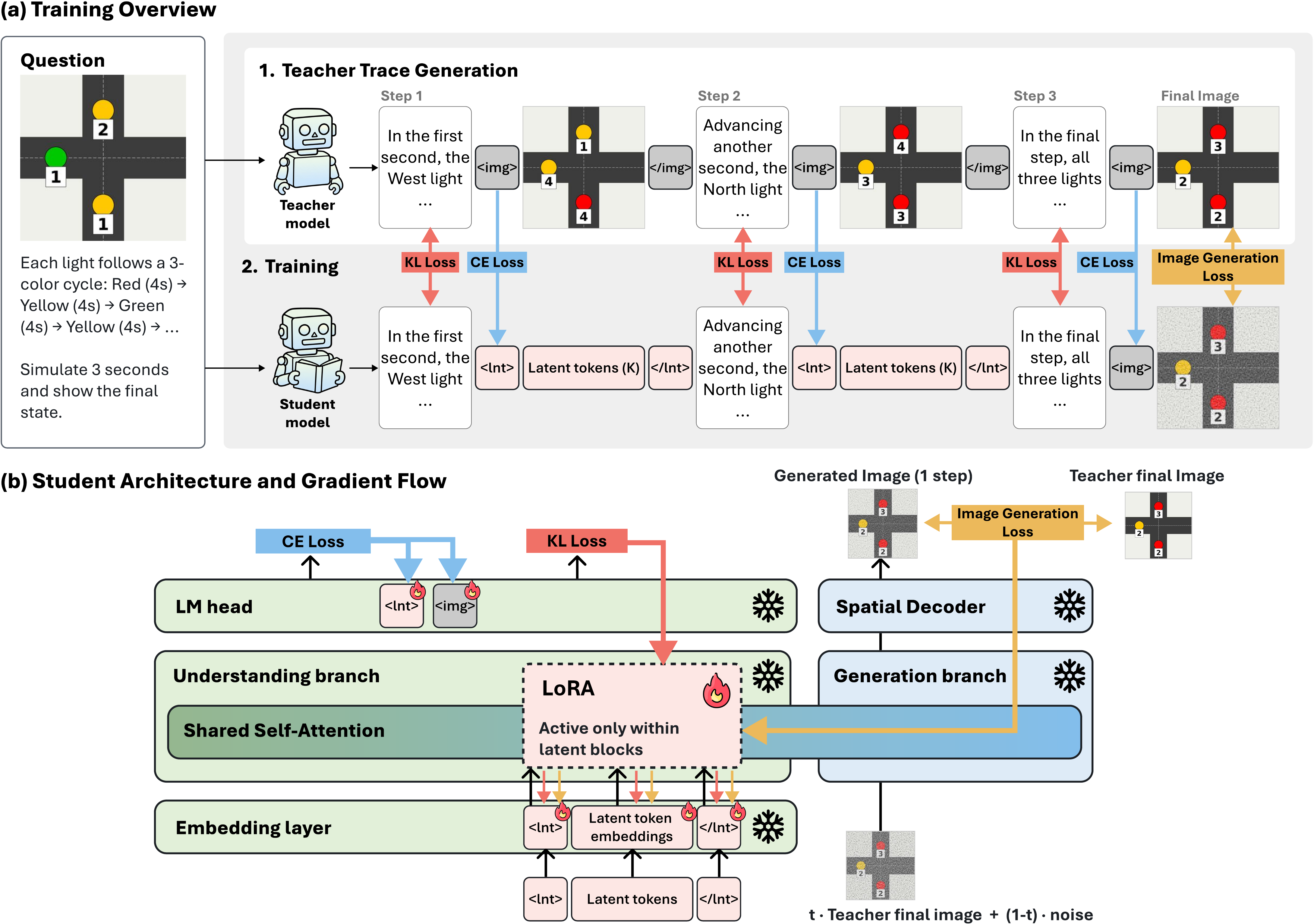}
    \label{training_overview}
    \caption{(a) Training Overview. The teacher generates interleaved traces, and the student is trained on the traces where intermediate images are replaced with latent blocks. (b) Student architecture and gradient flow. We train the latent token embeddings, latent block-specific LoRA, and the output embeddings for the \texttt{<lnt>} and \texttt{<img>} tokens, while keeping the remaining parameters frozen.}
    \label{fig:training_overview}
\end{figure*}

\subsection{Preliminaries}

\texttt{SenseNova-U1.5} employs a Mixture-of-Transformers architecture using shared self-attention, with modality-specific branches. The understanding branch processes multimodal inputs to generate text, and emitting an \texttt{<img\textgreater} token triggers the generation branch to produce an image.
During interleaved reasoning, the model alternates between text and intermediate images $v_i$ before producing the final answer image $a$. Because the understanding branch controls image generation, intermediate images can be replaced by other tokens without disrupting the final image generation.

\subsection{Interleaved Latent Reasoning of the Unified Model}

We aim to replace explicit intermediate images with compact latent representations that support subsequent reasoning and final-image generation. To achieve this, we design a latent reasoning mechanism where the intermediate explicit images $v_i$ are replaced by latent blocks $z_i$. Each latent block $z_i$ consists of $K$ learnable embeddings enclosed by special tokens \texttt{\textless lnt\textgreater} and \texttt{\textless /lnt\textgreater}.

During inference, instead of generating pixel-level images at each intermediate step, the understanding branch processes these latent tokens. These tokens help the subsequent reasoning to effectively reach the final answer $a$. The generation branch is used only for generating a final image $a$ when \texttt{\textless img\textgreater} is predicted by the understanding branch. The number of latent blocks and the timing of the final image are determined by the model.

\subsection{Offline Self-Distillation}

\paragraph{Teacher Trace Collection.} We use \texttt{SenseNova-U1.5} as a teacher model. Given a multimodal query $q = (q_{\text{text}}, q_{\text{img}})$, the teacher model generates interleaved text--image traces for offline self-distillation, as illustrated in \Cref{fig:training_overview} (a). During this process, we constrain the number of generated images to match the ground-truth image count. Finally, we use rejection sampling to keep only the traces whose final image passes the task-specific scorer.

\paragraph{Student Sequence Construction.} As illustrated in the lower part of \Cref{fig:training_overview} (a), we construct the student sequence by replacing each intermediate image in the teacher trace with a latent block consisting of $K$ latent tokens enclosed by \texttt{\textless lnt\textgreater} and \texttt{\textless /lnt\textgreater}, while keeping the reasoning text unchanged. We retain the final \texttt{\textless img\textgreater} token and use the teacher's final image as the flow matching target. We also add latent block usage instructions to the Oddin's system prompt (See \Cref{app:prompts_setting}).

\subsection{Training Objectives}
\label{sec:training_objectives}
We train the student model using image generation loss, text distillation loss, and special-token cross-entropy loss, where $\lambda$ denotes the weights and $T$ the distillation temperature:
\begin{equation}
    \mathcal{L}=\lambda_{\mathrm{gen}}\mathcal{L}_{\mathrm{gen}}+\lambda_{\mathrm{KL}}T^2\mathcal{L}_{\mathrm{KL}}+\lambda_{\mathrm{CE}}\mathcal{L}_{\mathrm{CE}}.
\end{equation}

\paragraph{Image generation loss.} We reconstruct the target image $x$ from a noisy state $x_t=tx+(1-t)s\epsilon$, where $\epsilon\sim\mathcal{N}(0,I)$, $t\in[0,1]$, and $s$ is the noise scale. Conditioned on the student's state $c_\theta$, the frozen generation branch predicts the clean image $\hat{x}_\theta=x_t+(1-t)v(x_t,t;c_\theta)$. Inspired by region-focused supervision \citep{LoVaRA}, we formulate the loss to explicitly penalize errors in edited areas:
\begin{equation}
    \mathcal{L}_{\mathrm{gen}}=\mathbb{E}_{t,\epsilon}\left[\operatorname{MSE}(\hat{x}_\theta,x)+\operatorname{MSE}_{M}(\hat{x}_\theta,x)\right],
\end{equation}
where $M$ is a binary mask indicating the changed regions between the input and target images. See \Cref{app:image-generation} for details.

\paragraph{Text distillation loss.} We align the output distributions of the image-conditioned teacher and the latent-conditioned student through teacher-to-student KL divergence:
\begin{equation}
    \mathcal{L}_{\mathrm{KL}}=\sum_{i\in\mathcal{P}_{\mathrm{text}}}\operatorname{KL}\left(p_{\mathrm{teacher},i}\,\middle\|\,p_{\mathrm{student},i}\right),
\end{equation}
where $\mathcal{P}_{\mathrm{text}}$ contains all aligned text-prediction positions, excluding predictions of \texttt{\textless lnt\textgreater} and \texttt{\textless img\textgreater}. Both distributions use temperature $T$ and retain the teacher's top-50 tokens, with the remaining probability mass aggregated into one tail probability. See \Cref{app:text-distillation} for details.

\paragraph{Special-token cross-entropy loss.} At teacher image-generation positions, we supervise \texttt{\textless lnt\textgreater} for intermediate latent blocks and \texttt{\textless img\textgreater} for the final image:
\begin{equation}
    \mathcal{L}_{\mathrm{CE}}=-\mathbb{E}_{i\sim\pi}\left[\log p_{\mathrm{student},i}(y_i)\right],
\end{equation}
where $y_i$ is the target special token and $\pi$ balances intermediate and final events. Probabilities are normalized over the full vocabulary. See \Cref{app:cross-entropy} for details.

\begin{table*}[t]
  \caption{Accuracy on held-out VBVR-Pro-RL testset. We report per-task accuracy and the average across all seven tasks. Best in \textbf{bold}, second best \underline{underlined}. The last row shows the difference of Oddin from Interleaved Reasoning.}
  \label{tab:stage1_native}
  \centering
  \footnotesize
  \setlength{\tabcolsep}{6pt}
  \renewcommand{\arraystretch}{1.1}
  \begin{tabular}{l ccccccc c}
    \toprule
    \textbf{Method}
      & \shortstack{Traffic\\Light}
      & \shortstack{Separate Object\\Spinning}
      & \shortstack{Dot to\\Dot}
      & \shortstack{Rotation\\Puzzle}
      & \shortstack{Track Object\\Movement}
      & \shortstack{Domino\\Chain}
      & \shortstack{Sliding\\Puzzle}
      & \textbf{Avg.} \\
    \midrule
    \multicolumn{9}{l}{\textit{Closed-source VLMs $+$ image generation tool}} \\
    \quad \texttt{GPT-4o} $+$
    \texttt{GPT-Image-2}                   & 0.22 & 0.08 & 0.02 & 0.05 & 0.42 & 0.00 & \underline{0.81} & 0.23 \\
    \quad \texttt{GPT-6.1-sol} $+$ \texttt{GPT-Image-2}                   & 0.45 & 0.56 & 0.02 & 0.21 & \textbf{0.86} & 0.00 & \textbf{0.99} & 0.44 \\
    \midrule
    \multicolumn{9}{l}{\textit{Unified Large Multimodal models (ULMMs)}} \\
    \quad \texttt{BAGEL-7B}                     & 0.00 & 0.00 & 0.00 & 0.00 & 0.01 & 0.00 & 0.00 & 0.00 \\
    \quad \texttt{ThinkMorph}                   & 0.00 & 0.00 & 0.00 & 0.00 & 0.02 & 0.00 & 0.00 & 0.00 \\
    \quad \texttt{SenseNova-U1.5} (Text CoT)    & 0.03 & 0.08 & 0.10 & 0.59 & \underline{0.61} & \underline{0.74} & 0.56 & 0.39 \\
    \quad \texttt{SenseNova-U1.5} (Interleaved) & \underline{0.52} & \underline{0.70} & \textbf{0.63} & \underline{0.64} & 0.37 & 0.53 & 0.25 & \underline{0.52} \\
    \midrule
    \rowcolor{oursblue}
    \textbf{Oddin} (ours)          & \textbf{0.68} & \textbf{0.80} & \underline{0.49} & \textbf{0.97} & 0.44 & \textbf{0.76} & 0.43 & \textbf{0.65} \\
    \rowcolor{lastrowbg}
    \quad $\Delta$ vs.\ Interleaved
      & \textcolor{deltagreen}{+0.16} & \textcolor{deltagreen}{+0.10} & \textcolor{deltared}{$-$0.14} & \textcolor{deltagreen}{+0.33}
      & \textcolor{deltagreen}{+0.07} & \textcolor{deltagreen}{+0.23} & \textcolor{deltagreen}{+0.18} & \textcolor{deltagreen}{+0.13} \\
    \bottomrule
  \end{tabular}
\end{table*}

\subsection{Trainable Parameters and Gradient Flow}
As shown in \Cref{fig:training_overview}(b), we train the latent tokens, the special tokens  \texttt{\textless lnt\textgreater} and \texttt{\textless /lnt\textgreater}, and LoRA adapters \cite{LoRA} while freezing all other parameters. LoRA is active only within latent blocks, including their special tokens. The image generation and text distillation losses backpropagate through frozen operations to the latent embeddings and adapters. For the special-token cross-entropy, we detach the hidden states to update only the two output embeddings. Initialization and adapter settings are provided in \Cref{app:trainable-parameters}.

\section{Experiment}
\subsection{Experimental Setup}
\paragraph{Training Data.} We use seven tasks from VBVR-Pro for which the teacher achieves over $20\%$ success under interleaved reasoning and the final image can be automatically verified. We train on $17{,}051$ successful teacher traces generated from 7K VBVR-Pro-RL and 21K native-resolution VBVR-Pro-SFT, using $1024 \times 1024$ images. Task selection and data construction are detailed in \Cref{app:experiment_details}.

\paragraph{Evaluation and Baselines.} We evaluate accuracy on $700$ held-out problems across seven VBVR-Pro-RL tasks, with $100$ problems per task. We compare our model against two closed VLMs (\texttt{GPT-4o} \citep{gpt4o} and \texttt{GPT-6.1-sol} \citep{gpt61sol}) with \texttt{GPT-Image-2} \citep{gptimages2}, and three ULMMs (\texttt{BAGEL-7B} \cite{BAGEL}, \texttt{ThinkMorph} \citep{ThinkMorph}, and \texttt{SenseNova-U1.5} \cite{SenseNova-U1.5}). For \texttt{SenseNova-U1.5}, we evaluate two modes: text CoT and interleaved generation. See \Cref{app:evaluation_details} for details.

\subsection{Result}
\paragraph{Main results.} Oddin achieves a mean accuracy of $0.65$, outperforming all baselines. The results show two distinct patterns depending on the relative performance of Interleaved Reasoning and text CoT. On tasks where Interleaved Reasoning outperforms text CoT (Traffic Light, Separate Object Spinning, Rotation Puzzle, and Dot to Dot), Oddin further improves performance without generating intermediate images. The only exception is Dot to Dot, where Oddin scores 0.14 points lower. On tasks where Interleaved Reasoning underperforms text CoT (Domino Chain, Track Object Movement, and Sliding Puzzle), Oddin consistently recovers performance. It achieves the highest score on Domino Chain, even surpassing text CoT. Moreover, Oddin reduces token generation by $84\%$ per problem compared to Interleaved Reasoning, replacing an average of $7.9$ images with just one final image and $4$ latent blocks (\Cref{app:inference_cost}).

\begin{figure}[t]
    \centering
    \includegraphics[width=\linewidth]{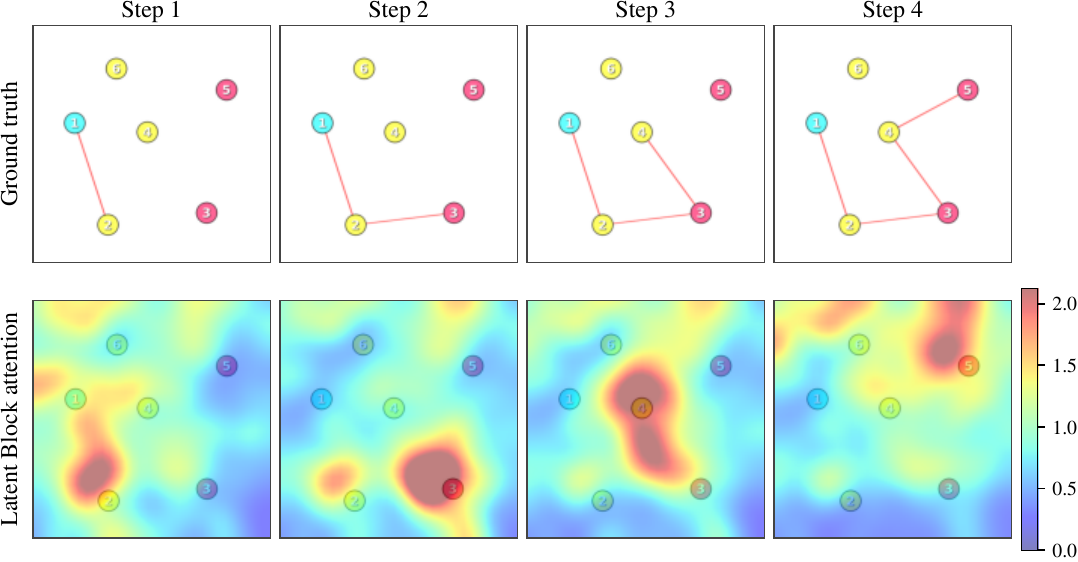}
    \caption{Gaussian smoothed attention of latent blocks on each step in Dot To Dot task. Values are relative to uniform attention over the image (1 = evenly spread).}
    \label{fig:qualitative_analysis}
\end{figure}

\paragraph{Qualitative analysis.}
Figure \ref{fig:qualitative_analysis} visualizes an example of the Gaussian-smoothed attention from each latent block at a corresponding intermediate step to the input image, averaged over its 4 latent tokens. In this Dot To Dot task, attention concentrates around the dot newly connected at that step, following the drawing order given in the task instruction. This suggests that the learned latent tokens carry compact, step-specific visual information aligned with the visual reasoning trace.

\section{Conclusion}
In this paper, we propose Oddin, a framework that enables ULMMs to perform visual reasoning using compact latent tokens processed by the understanding branch. Trained via offline self-distillation, Oddin learns the visual reasoning capabilities without explicit intermediate supervision and outperforms baselines on average. Qualitative analysis shows that the learned latent tokens attend to the relevant regions to each reasoning step, indicating that ULMMs can perform visual reasoning within the understanding branch.

\paragraph{Limitations and future work.} In some VBVR-Pro tasks, Oddin shows lower performance compared to the strongest baselines. This indicates that while latent representations maintain global semantic reasoning, they lack precise spatial information. Future work will focus on addressing detailed visual manipulation tasks within the latent space.

{
    \small
    \bibliographystyle{ieeenat_fullname}
    \bibliography{main}
}
\appendix
\crefname{section}{Appendix}{Appendices}
\Crefname{section}{Appendix}{Appendices}

\section{Training Objectives Details}
This section provides the detailed formulations and implementation specifics for the three loss components introduced in Section~\ref{sec:training_objectives}.

\subsection{Image Generation Loss}
\label[appendix]{app:image-generation}

The target $x$ is defined as the final image from a successful teacher trace. Images are normalized to $[-1,1]$. The noisy input is formulated as $x_t=tx+(1-t)s\epsilon$, where $s$ denotes the noise scale of the pretrained generation branch. The timestep $t$ is computed as $t=1-3u/(1+2u)$ with $u\sim\mathcal{U}(0,1)$. The velocity prediction yields the clean-image estimate $\hat{x}_\theta=x_t+(1-t)v(x_t,t;c_\theta)$.

The change mask is constructed by resizing the input image to the target resolution using bilinear interpolation. A pixel is selected if its absolute difference from the target exceeds $0.1$ in any RGB channel. A $9\times9$ binary dilation is applied to incorporate neighboring pixels. Let $e_{chw}=\hat{x}_{\theta,chw}-x_{chw}$ denote the prediction error. The loss terms are defined as:
\begin{align}
    \operatorname{MSE}(\hat{x}_\theta,x)
    &=\frac{1}{3HW}\sum_{c,h,w}e_{chw}^2,\\
    \operatorname{MSE}_{M}(\hat{x}_\theta,x)
    &=\frac{\sum_{c,h,w}M_{hw}e_{chw}^2}{\max\left(3\sum_{h,w}M_{hw},1\right)}.
\end{align}
The masked term averages errors across the selected pixels and RGB channels, evaluating to zero for empty masks.

\subsection{Text Distillation Loss}
\label[appendix]{app:text-distillation}

Both models process identical reasoning text via teacher forcing. The teacher receives intermediate images, whereas the student receives latent blocks. Positions are aligned by text tokens, including text preceding the first latent block, excluding prediction positions for \texttt{\textless lnt\textgreater} or \texttt{\textless img\textgreater}. Teacher distributions are computed without gradients using the base model and system prompt.

Let $q_{A,i}$ denote the full-vocabulary distribution for $A\in\{\mathrm{teacher},\mathrm{student}\}$, and $K_i$ denote the teacher's top-50 tokens. To formulate the KL objective, we define the aggregated distribution $p_{A,i}$ over the modified vocabulary $K_i \cup \{\text{tail}\}$ as:
\begin{equation}
    p_{A,i}(w) = 
    \begin{cases} 
        q_{A,i}(w), & \text{if } w \in K_i, \\
        1 - \sum_{v \in K_i} q_{A,i}(v), & \text{if } w = \text{tail}.
    \end{cases}
\end{equation}
This formulation extracts 50 token categories and one aggregated tail category. The teacher-to-student KL divergence is summed across text positions per trace. The total objective is scaled by a $T^2$ multiplier.

\subsection{Special-token Cross-Entropy Loss}
\label[appendix]{app:cross-entropy}

At positions preceding teacher image-generation events, target tokens are \texttt{\textless lnt\textgreater} for intermediate images and \texttt{\textless img\textgreater} for the final image. Cross-entropy is calculated over the full student vocabulary without tail aggregation used in \Cref{app:text-distillation}. Within each trace, intermediate \texttt{\textless lnt\textgreater} and final \texttt{\textless img\textgreater} losses are averaged independently and weighted equally. Traces lacking intermediate images optimize only the final \texttt{\textless img\textgreater} loss. This weighting schema defines the parameter $\pi$. The preceding hidden states are detached to restrict gradients strictly to the two special-token output rows.

\subsection{Trainable Parameters}
\label[appendix]{app:trainable-parameters}

The input embeddings of \texttt{\textless lnt\textgreater} and \texttt{\textless /lnt\textgreater} are initialized from the pretrained \texttt{\textless img\textgreater} and \texttt{\textless /img\textgreater} embeddings, respectively. Each latent token embedding is initialized from \texttt{\textless IMG\_CONTEXT\textgreater}. Latent token embeddings are independently trainable, and the same ordered set is reused across all blocks. 

LoRA adapters are active exclusively within latent blocks, including their boundary tokens. The two special-token output rows are initialized independently from the pretrained \texttt{\textless img\textgreater} output row and remain separate from the input embeddings. The closing token \texttt{\textless /lnt\textgreater} is inserted after $K$ latent tokens.

All other parameters remain frozen. The image generation and text distillation losses update the latent embeddings and LoRA adapters. The text distillation loss additionally updates the two special-token output rows. The special-token cross-entropy loss updates only these output rows due to its detached-hidden-state computation.

\section{Experiment Details}
\label[appendix]{app:experiment_details}
\subsection{Hyperparameters}
\label[appendix]{app:hyperparameters}
The main configuration uses $K=4$ latent tokens per latent block. LoRA adapters are configured with rank 16 and $\alpha=32$. The model is trained for two epochs with a global batch size of 16. The learning rate is initialized at $2\times10^{-4}$ and follows a cosine decay schedule. The loss coefficients are set to $\lambda_{\mathrm{gen}}=10$, $\lambda_{\mathrm{KL}}=0.005$, $\lambda_{\mathrm{CE}}=0.1$, and the temperature is $T=1.5$.

\subsection{Training Data}
\label[appendix]{app:training_data} 
\paragraph{Task Selection.} We select tasks from VBVR-Pro using two criteria: (1) the teacher achieves a task success rate above $20\%$ under interleaved reasoning, ensuring that success traces for self-distillation can be collected and (2) task success can be automatically verified from the final generated image. This yields seven tasks: Traffic Light (O-52), Separate Object Spinning (G-25), Dot to Dot Task (O-34), Rotation Puzzle (O-44), Track Object Movement (G-8), Domino Chain Branch Path Prediction (O-23), and Sliding Puzzle (O-47). For brevity, we refer to the Domino Chain Branch Path Prediction task as Domino Chain.

\paragraph{Data Construction.} The VBVR-Pro-RL pool contains $7{,}000$ problems across the seven selected tasks. We hold out $700$ problems for evaluation and use the remaining 6,300 as training candidates, together with 21,000 randomly sampled VBVR-Pro-SFT problems. VBVR-Pro-SFT input images are rendered directly at $1024\times1024$ resolution from their rendering specifications. We retain teacher traces whose final generated image passes task verification, resulting in 17,051 training traces. Our grader is adapted from the official VBVR-Pro evaluation code. The image reconstruction target is the verified teacher-generated final image.

\subsection{Evaluation Details}
\label[appendix]{app:evaluation_details}
\paragraph{Baselines.} We compare our method against two groups of baselines. First, for closed-source VLMs, we evaluate \texttt{gpt-4o} and \texttt{gpt-6.1-sol}. Since they cannot generate images natively, we pair them with \texttt{GPT-Image-2} to produce the final image based on their textual reasoning. Second, for open-weight ULMMs, we evaluate \texttt{BAGEL-7B}, \texttt{ThinkMorph}, and \texttt{SenseNova-U1.5}. Specifically for \texttt{SenseNova-U1.5}, we compare against two inference modes: Text CoT, which performs textual reasoning and generates only the final image, and Interleaved reasoning, which generates explicit intermediate images alongside text. Our model replaces these intermediate images with latent blocks while retaining final-image generation.

\paragraph{Evaluation configurations.}
All open-weight models use greedy text decoding with maximum $4{,}096$ new text tokens, $1024^2$ output resolution, $50$ denoising steps, and text CFG scale $4.0$. The remaining sampler settings follow each model's official defaults. \texttt{SenseNova-U1.5} and Oddin use image CFG $1.0$ without CFG renormalization, whereas \texttt{BAGEL-7B} and \texttt{ThinkMorph} use image CFG $2.0$ with CFG renormalization. For closed-source VLMs, we make one call per problem through the OpenAI Responses API, with the input frame and the task prompt. The model reasons freely and draws its answer by calling the \texttt{GPT-Image-2} image generation tool ($1024\times1024$, medium effort). The number of tool calls is left to the model, and the maximum output length is $32{,}000$ tokens. All other decoding settings are the API defaults, and denoising steps and CFG are not controllable.

\paragraph{Generation and Answer Selection.} During inference, except \texttt{BAGEL-7B} no method receives the ground-truth image count. \texttt{BAGEL-7B} is designed to generate one image per call and is therefore called $n_{\mathrm{gt}}$ times. Our model autonomously selects whether to emit another latent block or generate the final image, with a maximum of 12 latent blocks. Interleaved Reasoning generates images without a prescribed count, subject to a maximum of 13 images. We score its last image before \texttt{\textless/think>} to exclude images redrawn in the subsequent answer. If there is no delimiter, we use the last generated image. For Text CoT and our model, we score the final generated image.

\begin{figure*}[!t]
  \centering
  \includegraphics[width=0.9\textwidth]{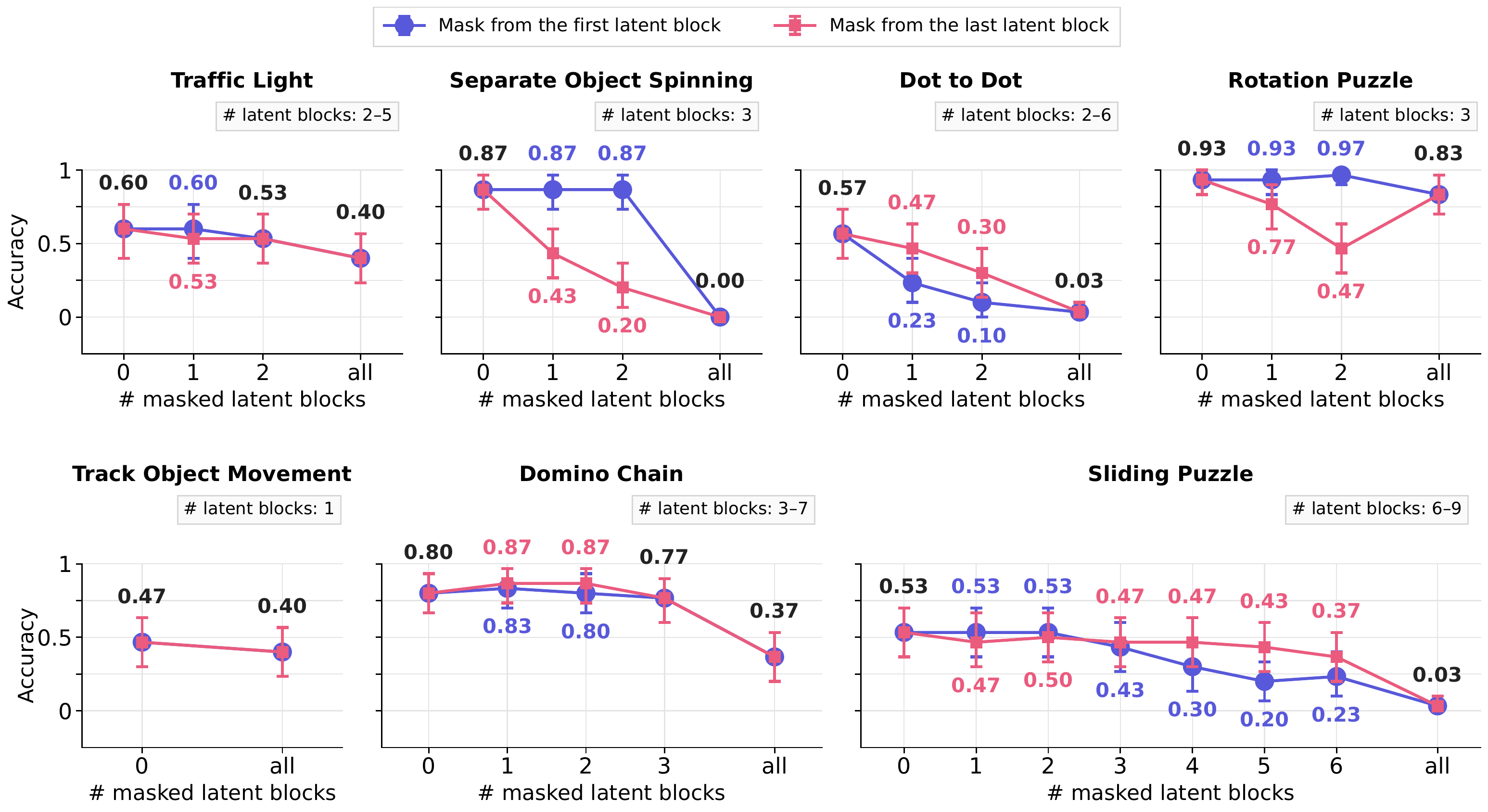}
  \caption{Masking analysis of latent blocks. Accuracy when latent blocks are masked from the first block (blue) or from the last block (red). ``all'' means masking every block. Black numbers mean identical value. Bars are 95\% bootstrap confidence intervals over 30 problems per task. The box gives the number of latent blocks per problem.}
  \label{fig:masking_blocks}
\end{figure*}

\subsection{Prompts Setting}
\label[appendix]{app:prompts_setting}

\newcommand{\commonsent}{Only modify what the task requires; keep the background and any other element unchanged.}

\paragraph{System prompts.}
Each model uses its own system prompt. \texttt{BAGEL-7B} generates one image per call, so we call it $n_{\mathrm{gt}}$ times and feed each generated image back as context.

\begin{promptbox}
\textbf{Oddin.} The SenseNova-U1.5 system message, followed by: \textit{In Think Mode, use \texttt{\textless latent\textgreater} blocks instead of visible images for intermediate visual reasoning. Generate a visible image only for the final requested state. When following the task's image-sequence instructions, count each latent block as one intermediate output image and the final generated image as the last output image.}
\end{promptbox}

\begin{promptbox}
\textbf{SenseNova-U1.5.} {\itshape You are a multimodal assistant capable of reasoning with both text and images. You support two modes:\par
Think Mode: When reasoning is needed, you MUST start with a \texttt{\textless think\textgreater\textless /think\textgreater} block and place all reasoning inside it. You MUST interleave text with generated images using tags like \texttt{\textless image1\textgreater}, \texttt{\textless image2\textgreater}. Images can ONLY be generated between \texttt{\textless think\textgreater} and \texttt{\textless /think\textgreater}, and may be referenced in the final answer.\par
Non-Think Mode: When no reasoning is needed, directly provide the answer without reasoning. Do not use tags like \texttt{\textless image1\textgreater}, \texttt{\textless image2\textgreater}; present any images naturally alongside the text.\par
After the think block, always provide a concise, user-facing final answer. Match the user's language in both reasoning and the final answer.}
\end{promptbox}

\begin{promptbox}
\textbf{BAGEL-7B.} \textit{You should first think about the planning process in the mind and then generate the image. The planning process is enclosed within \texttt{\textless think\textgreater} \texttt{\textless /think\textgreater} tags, i.e. \texttt{\textless think\textgreater} planning process here \texttt{\textless /think\textgreater} image here}
\end{promptbox}

\begin{promptbox}
\textbf{ThinkMorph-7B.} \textit{Let's think step by step to answer the question. For text-based thinking, enclose the process within \texttt{\textless think\textgreater} \texttt{\textless /think\textgreater}. For visual thinking, enclose the content within \texttt{\textless image\_start\textgreater} \texttt{\textless /image\_end\textgreater}. Finally conclude with the final answer wrapped in \texttt{\textless answer\textgreater} \texttt{\textless /answer\textgreater}}
\end{promptbox}

\begin{promptbox}
\textbf{GPT-4o, GPT-6.1-sol.} \textit{You are a multimodal assistant capable of reasoning with both text and images. Reason about the task in whatever way you find best. After the reasoning, always provide a concise, user-facing final answer. The final answer should be an image showing the final state of the scene once the task is complete, generated with the image generation tool. Match the user's language in both reasoning and the final answer.}
\end{promptbox}

\paragraph{Task prompts.}
All open-source models receive the original VBVR-Pro task prompt without modification, as \texttt{[image, prompt]}. Since the closed-source VLMs reason freely on how to solve each task, they receive the same prompt without its final sentence, which prescribes the number of output images, and with "Provide an image showing the final state." appended.

\section{Additional Analyses}

\subsection{Additional Qualitative Analysis}
\begin{figure}[H]
  \centering
  \includegraphics[width=\linewidth]{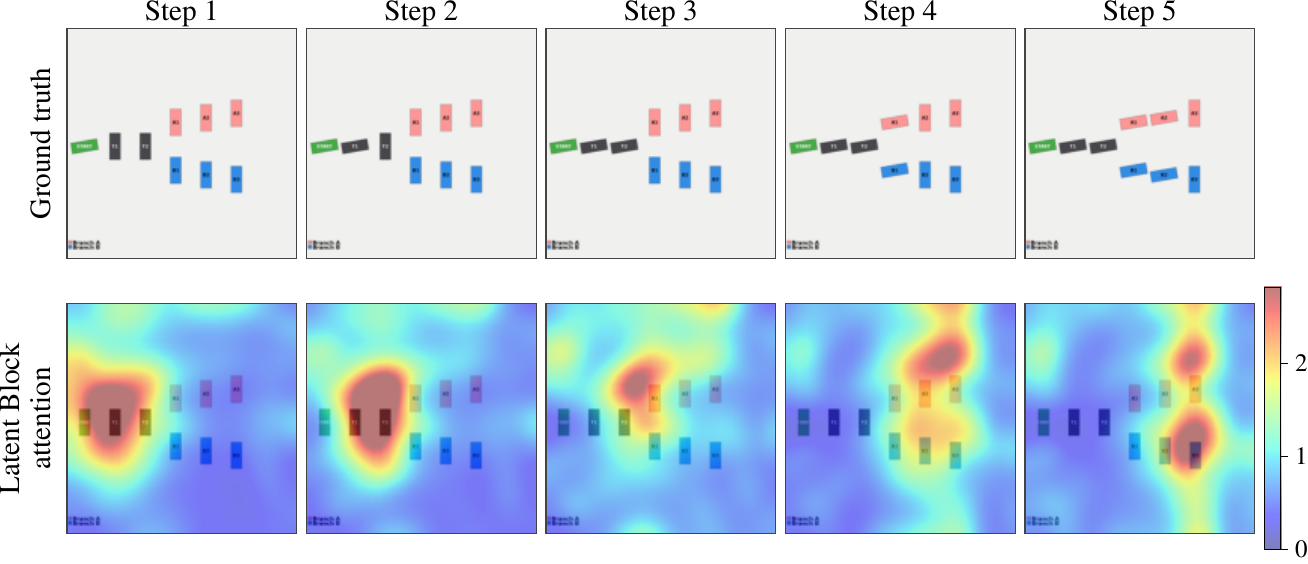}
  \caption{Gaussian-smoothed attention of the latent block at each step on Domino Chain Branch Path Prediction, shown as in Figure \ref{fig:qualitative_analysis}. Attention concentrates around the dominoes that fall at the current and the next step.}
  \label{fig:additional_qualitative_analysis}
\end{figure}

\Cref{fig:additional_qualitative_analysis} shows the latent-block attention on a Domino Chain Branch Path Prediction problem, visualized as in \Cref{fig:qualitative_analysis}. As the chain propagates, attention moves along with it, from the starting dominoes to the two branches. Unlike Dot To Dot, attention often runs one step ahead, concentrating on the dominoes about to fall at the next step. This suggests that the latent blocks anticipate the upcoming state transition rather than only encoding the current one.

\subsection{Masking Analysis of Latent Blocks}
\label[appendix]{app:masking}

To test whether the latent blocks are used for visual reasoning, we mask them from the attention at the final image generation step. In \Cref{fig:masking_blocks}, both settings mask up to the minimum number of latent blocks present in each task, while the ``all'' condition masks every block. Masking every block lowers accuracy in most tasks and brings it close to zero in Separate Object Spinning, Dot to Dot, and Sliding Puzzle. Moreover, in Separate Object Spinning, masking from the last block reduces the accuracy more than masking from the first, indicating that the final image generation relies more on the later blocks. In Dot to Dot, the opposite holds: masking from the first block reduces the accuracy more, indicating that earlier blocks carry information the later ones cannot replace. Overall, the latent blocks affect the generation of the final image, and the blocks that matter differ across tasks.

\subsection{Inference Cost}
\label[appendix]{app:inference_cost}

\Cref{tab:token_cost} reports the number of tokens generated per problem, counted separately for text, latent, and image tokens. Interleaved reasoning generates on average $7.9$ images ($8{,}135$ image tokens), whereas Oddin generates a single final image and about $4$ latent blocks ($24$ latent tokens), reducing the total number of generated tokens by $84\%$. Oddin also generates $49\%$ fewer text tokens than Interleaved reasoning ($725 \rightarrow 371$). Its total is close to that of Text CoT ($1{,}209$).

\begin{table}[t]
  \caption{Average number of generated tokens per problem on the 700 held-out problems. Text tokens include special tokens in the generated text (e.g., \texttt{<think>}). Each $1024\times1024$ image counts as $1{,}026$ tokens ($1{,}024$ image tokens plus \texttt{<img>} and \texttt{</img>}). A latent block counts as $6$ tokens (\texttt{<lnt>}, $K{=}4$ slots, \texttt{</lnt>}). Input tokens are excluded.}
  \label{tab:token_cost}
  \centering
  \footnotesize
  \setlength{\tabcolsep}{4pt}
  \renewcommand{\arraystretch}{1.1}
  \begin{tabular}{l rrrr}
    \toprule
    \textbf{Method} & Text & Latent & Image & \textbf{Total} \\
    \midrule
    Text CoT     & 183 & --  & 1,026 (1 img) & 1,209 \\
    Interleaved  & 725 & --  & 8,135 (7.9 imgs) & 8,860 \\
    \rowcolor{oursblue}
    \textbf{Oddin} (ours) & 371 & 24 (4 blocks) & 1,026 (1 img) & \textbf{1,421} \\
    \bottomrule
  \end{tabular}
\end{table}

\end{document}